\documentclass[11pt]{article}
\usepackage[table]{xcolor}
\usepackage{geometry}
\usepackage{coling2020}
\usepackage{times}
\usepackage{url}
\usepackage{latexsym}
\usepackage{microtype}
\usepackage{multirow}
\usepackage{booktabs}
\usepackage[normalem]{ulem}
\usepackage[utf8]{inputenc}
\usepackage{fourier} 
\usepackage{array}
\usepackage{makecell}
\usepackage{amssymb}
\def\chk{\checkmark}
\usepackage{graphicx}
\usepackage{subcaption}
\usepackage{mathptmx}
\usepackage{fancyhdr}

\DeclareMathSymbol{\ast}{\mathbin}{symbols}{"03}
\newcommand{\STAB}[1]{\begin{tabular}{@{}c@{}}#1\end{tabular}}

\colingfinalcopy 

\newcommand{\ea}{$^\dagger$}
\newcommand{\eb}{$^\ddagger$}

\title{GUIR at SemEval-2020 Task 12: \\ Domain-Tuned Contextualized Models for Offensive Language Detection}

\author{Sajad Sotudeh\thanks{~~~equal contribution}~~\ea, Tong Xiang\footnotemark[1]~~\eb, Hao-Ren Yao\ea, Sean MacAvaney\ea, \\ \bf Eugene Yang\ea, Nazli Goharian\ea, Ophir Frieder\ea \\
  Information Retrieval Lab \\
  Georgetown University, USA \\
  \ea {\tt \{firstname\}@ir.cs.georgetown.edu}\\
  \eb {\tt tx39@georgetown.edu}
}

\date{}

\begin{document}
\maketitle
\begin{abstract}
Offensive language detection is an important and challenging task in natural language processing.
We present our submissions to the OffensEval 2020 shared task, which includes three English sub-tasks: identifying the presence of offensive language (Sub-task A), identifying the presence of target in offensive language (Sub-task B), and identifying the categories of the target (Sub-task C).
Our experiments explore using a domain-tuned contextualized language model (namely, BERT) for this task. We also experiment with different components and configurations (e.g., a multi-view SVM) stacked upon BERT models for specific sub-tasks. Our submissions achieve F1 scores of 91.7\% in Sub-task A, 66.5\% in Sub-task B, and 63.2\% in Sub-task C. 
We perform an ablation study which reveals that domain tuning considerably improves the classification performance.
Furthermore, error analysis shows common misclassification errors made by our model and outlines research directions for future. 
\end{abstract}
\section{Introduction}
\blfootnote{This work is licensed under a Creative Commons Attribution 4.0 International License. License details: \url{https://creativecommons.org/licenses/by/4.0/}.}
The rapid development of user-generated content in social media has given millions of people the ability to easily share their ideas with each other. While users can publicly communicate their beliefs with others, their published content may be offensive to other individuals or groups. Since offensive speech can jeopardize others' ability to express themselves, many social media platforms
restrict the type of acceptable content on their platforms. Manually detecting such content is expensive and time-consuming. Thus, automatic detection of these behaviors in social media has attracted researchers' attention.
Although this topic has been explored in prior works~\cite{10.1145/3308560.3314196,Rizos2019AugmentTP,Zhang2018DetectingHS,Nobata2016AbusiveLD}, the task of offensive language detection still remains challenging.
The OffensEval 2020 shared task~\cite{zampieri-etal-2020-semeval} aims to encourage continued work in this area, and we progress the study of offensive language detection through our participation in this shared task utilizing contextualized language models.

OffensEval 2020 evaluates various aspects of offensive language following the scheme of the OLID dataset~\cite{Zampieri2019PredictingTT}, including 
identifying the presence of offensive language~(Sub-task A), identifying whether the offensive language is targeted~(Sub-task B), and identifying whether the target of the offensive language is an individual,  group,  or something else~(Sub-task C).
The task extends the prior work by introducing a new data collection from Twitter that spans multiple languages. 

Our experiments explore various versions of the BERT model (Bidirectional Encoder Representations from Transformers \cite{Devlin2019BERTPO}) tuned for offensive language identification. As previous studies have mentioned, dealing with social media content can be challenging because it is often short and noisy~\cite{Qian2018LeveragingIA}. To alleviate this problem, we first fine-tune a BERT model on a large amount of unlabelled data gathered from Twitter. We pre-process the tweets by splitting hashtags and replacing emoji with textual descriptions. We also explore an ensemble learning method (i.e., a multi-view SVM model akin to~\cite{macavaney2019hate}) which combines different word n-gram features of the input text as views and predict the output based on the combination of these views. We report competitive performance of these models, which achieve macro F1 scores of 91.7\%, 66.5\%, and 63.2\% on Sub-tasks A, B and, C, ranking 5th, 6th, and 11th for each sub-task out of 85, 43, 39 participated teams respectively.
    
Our contributions are as follows: 1) 
we present variations of BERT model tuned for different aspects of offensive language detection; 2) 
we report competitive results of our models with detailed comparisons; 3) we perform an ablation study to examine the effect of different components in the proposed systems; and 4) we conduct an error analysis to provide insights into how our models perform the classification, and where to focus in future work.
\section{Related work}\label{sec:background}

The research of automatic offensive language detection
has gained attention in the past decade. Most prior work in the domain employs the supervised learning paradigm~\cite{schmidt2017survey}.
Traditional models often made use of rule-based methods, such as template-based strategy~\cite{warner2012detecting} or pre-defined black-lists~\cite{xiang2012detecting}. Aside of methods, a widely used category of features are surface-level features, e.g., n-gram features~\cite{schmidt2017survey}. These features are often highly predictive, and can be easily combined with other approaches. Others explored the combination of n-gram features with part-of-speech~\cite{Nobata2016AbusiveLD} as well as dependency information~\cite{chen2012detecting} for offensive language detection. These approaches leverage prior linguistic knowledge in order to generate features. However, the generated features are usually derived from pre-existing natural language processing systems, which could lead to the propagation of errors in the models~\cite{zeng2014relation}. These features are usually combined with classical machine learning classifiers such as SVM~\cite{Yin2009DetectionOH,malmasi2018challenges} and Logistic Regression~\cite{Waseem2016HatefulSO,davidson2017automated}. Others have explored using a multi-view learning paradigm~\cite{zhao2017multi} with an ensemble classifier ~\cite{macavaney2019hate}.


Deep learning techniques have also been shown to be effective for offensive language detection~\cite{Badjatiya2017DeepLF,Park2017OnestepAT,Liu2019FuzzyML}, with approaches such as Convolution Neural Networks (CNN)~\cite{gamback2017using} and Long Short-Term Memory (LSTM) Networks~\cite{pitsilis2018detecting}. More recently, pre-trained transformer-based networks, such as BERT~\cite{Devlin2019BERTPO} have shown great advantages in learning context-sensitive word representations. In OffensEval 2019, BERT-based and ensemble methods were the most effective approaches~\cite{Zampieri2019SemEval2019T6,Liu2019nuli,Han2019jhan014AS}.
\section{Methodology}
\label{sec:method}
In our experiments, we utilize a pre-trained contextualized language model, namely BERT~\cite{Devlin2019BERTPO}, to identify the offensive language. Also, we explore techniques for pre-processing, contextualized language modeling, and model outputs ensembling.


\textbf{Pre-processing.} Given the unique conventions of language on Twitter, we explore several tokenization pre-processing techniques to enable the downstream models to encode information more effectively. Since the length of a tweet is limited to 280 characters, emoji are often used to convey emotions and tones efficiently. To address differences in user preferences among users, we replace emoji with a textual description of the icons in order to shorten the domain gap between the corpus and the tweets.
We use the the mapping of emoji to English descriptions from the open source Python package \texttt{emoji}.\footnote{\url{https://github.com/carpedm20/emoji}} 

Hashtags are another common convention in tweets. They are used to describe topics related to certain tweets. They often consist of several words concatenated together, such as \textit{\#VoteRedSaveAmerica} and \textit{\#trumptrain}. Since hashtags do not contain whitespaces at word boundaries, additional logic is required for segmentation. We utilize the open source \texttt{wordsegment}~\footnote{\url{https://github.com/grantjenks/python-wordsegment}} library to obtain the boundaries and further construct the original textual tokens.

The OffensEval 2020 dataset replaces all usernames with a \textit{@USER} placeholder. This can result in some long strings of redundant and repetitive placeholders because tweets are often prefixed with numerous users. We tokenize using the \texttt{nltk} tweet tokenizer~\cite{bird2009natural} and drop \textit{@USER} tokens if repeated more than three times consecutively to avoid redundant information (similar to~\newcite{Liu2019nuli}). Furthermore, the token \textit{URL}, which is the artificial placeholder for any URL encountered in tweets, is also replaced with \textit{http} to match the vocabulary in the pre-trained embeddings.


\textbf{Contextualized language modeling.} We utilize the BERT contextualized language model~\cite{Devlin2019BERTPO}. Since there are language differences between the formal text that BERT is trained on (i.e., Wikipedia and books) and social media posts (i.e., tweets), we first tune the model to the particular domain, akin to the domain pre-training approach described in~\cite{gururangan2020-dont-stop}. This is accomplished by taking the original model and continuing to train the masked language model and next sentence prediction objectives using a large amount of unlabeled tweets. Note that we do not extend the vocabulary; we rely on the model's original WordPiece tokens.


We then fine-tune the model for identifying offensive language utilizing labeled training data. Since the task is sequence classification, we utilize the classification mechanism of the model (i.e., a linear layer on top of BERT's classification token). We train the model minimizing the cross-entropy loss, as compared to the gold training labels. We train a separate model for each sub-task during experiments. 



\textbf{Model outputs ensembling.} As mentioned in Section~\ref{sec:background}, ensemble approaches are often beneficial for offensive language detection and related tasks. In this work, we extend the multi-view SVM approach from~\cite{macavaney2019hate} with the addition of features from the contextualized language model classifier. Specifically, linear SVM classifiers (view-classifiers) using various n-gram ranges~\footnote{While we experimented with different ranges of n-grams, 6-gram feature was the optimal one so we fixed n-gram at 6.} are first trained for each sub-task in addition to the BERT-based classifier. Then, the outputs of the view-classifiers (probability output from SVM and sigmoid output from BERT) are concatenated as a feature vector for a final linear SVM classifier (the meta-classifier). For the SVM view-classifiers, we explore using both L1 and L2 regularization.



\section{Experiment}

In this section, we present settings, results and analysis for our experiments.
We first give a brief introduction of the dataset used for training and evaluation. Then we show our experimental settings, and perform a comprehensive analysis including experimental results analysis, ablation analysis, and error analysis\footnote{Our error analysis contains tweet examples and words that are offensive in nature.} over our models' results. 

\subsection{Data}


\newcite{Zampieri2019PredictingTT} introduced the Offensive Language Identification Dataset (OLID), a large-scale dataset of English tweets constructed by searching for specific keywords that may include offensive words on Twitter. They developed a hierarchical annotation schema that determines: 1) if the tweet is offensive (OFF) or non-offensive (NOT); 2) if an OFF tweet is targeted (TIN), or untargeted (UNT); and 3) if a TIN tweet is targeted toward individual (IND), group (GRP), or others (OTH). 
We refer the readers to~\newcite{Zampieri2019PredictingTT} for more details about the dataset characteristics. 

The OffensEval 2020 task~\cite{zampieri-etal-2020-semeval} offered a multilingual offensive language detection dataset. We participate in the three sub-tasks under the English track. For this track, the training set from OLID is used as training data and the test set from OLID is treated as development data. The annotation for newly annotated test data follows the same hierarchical schema as OLID, which was used during the evaluation phase.
The task also provides a distant dataset~\cite{rosenthal2020largescale} including over $9M$ tweets with predicated labels from an ensemble of classifiers. For our experiments, we disregard the labels and only make use of the text as pre-training data for Twitter domain adaptation.





\subsection{Experimental settings}

For domain pre-training (as described in Section~\ref{sec:method}), we utilize the tweets from the distant dataset provided by the shared task (disregarding labels). We tune the \texttt{BERT-Base} model using these data via training on the language modeling task and the default hyper-parameters provided by the BERT authors for training (learning rate: $2\times10^{-5}$, masking rate: 0.15, maximum sequence length: 128). For better reproducibility, we use the authors' original implementation for this tuning.\footnote{\url{https://github.com/google-research/bert}}

To tune the BERT model for the specific task, we utilize the OLID training data. The input sentences are directly tokenized into subword units by the BERT WordPiece tokenizer. Additionally, each input sentence is concatenated with a special token \texttt{[CLS]} at the beginning.
Since tweets have a character length limitation, we define the maximum sequence length to be 256 tokens. 
Some of our experiments also make use of additional tokenization pre-processing techniques described in Section~\ref{sec:method}.  For the task tuning, we utilize the \texttt{transformers} library~\cite{Wolf2019HuggingFacesTS}.

We tune hyperparameters based on the F1 score on the development set using two approaches. First, we try a simple approach in which the development F1 performance is evaluated after each training epoch (1 to 10). Second, we explore utilizing the training loss as an early stopping signal. Once the loss value reaches a pre-defined range, 
the model is evaluated on the development set. 

We ensemble n-gram SVM view classifiers with the BERT models. The meta classifier consumes the probabilistic prediction from each view classifier to provide the final prediction. L1 and L2 regularization strategies are applied to all SVM models with the inverse regularization penalty $C$ to be $10^{-5}$.
 



\subsection{Results and discussion}

\begin{table}
\centering\small



\scalebox{0.9}{
\begin{tabular}{lccccrr}
\multicolumn{7}{c}{(a) Sub-task A} \\
\toprule
Model & \textsc{Tkn} & \textsc{Twd} & \textsc{Lt} & mSVM & Dev (F1) & Test (F1) \\
\midrule
BERT & - & - & - & - & 0.805 & 0.915 \\
BERT & \chk & - & - & - & 0.806 & 0.904 \\
BERT & \chk & \chk & - & L1 & 0.822 & 0.915 \\
BERT & \chk & \chk & - & - & 0.823 & 0.911 \\
BERT & - & \chk & \chk & - & \textbf{0.828} & * \textbf{0.917} \\
\midrule
\multicolumn{5}{l}{Top OffensEval 2019 / 2020} & 0.829 & 0.922 \\
\multicolumn{5}{l}{Median OffensEval 2019 / 2020} & 0.739 & 0.909 \\
\multicolumn{5}{l}{Mean OffensEval 2019 / 2020} & - & 0.871 \\
\multicolumn{5}{l}{Min OffensEval 2019 / 2020} & 0.171 & 0.073 \\
\multicolumn{5}{l}{Std OffensEval 2019 / 2020} & - & 0.127 \\
\bottomrule
\end{tabular}

\hspace{1em}
\begin{tabular}{lccccrr}
\multicolumn{7}{c}{(b) Sub-task B} \\
\toprule

Model & \textsc{Tkn} & \textsc{Twd} & \textsc{Lt} & mSVM & Dev (F1) & Test (F1) \\
\midrule
BERT & \chk & - & - & - & 0.718 & 0.651 \\
BERT & - & - & - & - & 0.745 & 0.287 \\
BERT & - & \chk & \chk & - & 0.761 & \textbf{0.701} \\
BERT & \chk & \chk & - & - & 0.815 & 0.648 \\
BERT & \chk & \chk & - & L2 & \textbf{0.843} & * 0.665 \\
\midrule
\multicolumn{5}{l}{Top OffensEval 2019 / 2020} & 0.755 & 0.746 \\
\multicolumn{5}{l}{Median OffensEval 2019 / 2020} & 0.638 & 0.569 \\
\multicolumn{5}{l}{Mean OffensEval 2019 / 2020} & - & 0.555 \\
\multicolumn{5}{l}{Min OffensEval 2019 / 2020} & 0.121 & 0.278 \\
\multicolumn{5}{l}{Std OffensEval 2019 / 2020} & - & 0.120 \\
\bottomrule
\end{tabular}
}

\vspace{1em}

\scalebox{0.90}{
\begin{tabular}{lccccrr}
\multicolumn{7}{c}{(c) Sub-task C} \\
\toprule
Model & \textsc{Tkn} & \textsc{Twd} & \textsc{Lt} & mSVM & Dev (F1) & Test (F1) \\
\midrule
mSVM & - & - & - & L1 & 0.488 & 0.388 \\
BERT & - & - & - & - & 0.502 & 0.536 \\
BERT & \chk & \chk & - & L2 & 0.602 & \textbf{0.654} \\
BERT & \chk & \chk & - & L1 & 0.625 & 0.649 \\
BERT & \chk & \chk & - & - & \textbf{0.631} & * 0.632 \\
\midrule
\multicolumn{5}{l}{Top OffensEval 2019 / 2020} & 0.660 & 0.715 \\
\multicolumn{5}{l}{Median OffensEval 2019 / 2020} & 0.515 & 0.581 \\
\multicolumn{5}{l}{Mean OffensEval 2019 / 2020} & - & 0.560 \\
\multicolumn{5}{l}{Min OffensEval 2019 / 2020} & 0.090 & 0.057 \\
\multicolumn{5}{l}{Std OffensEval 2019 / 2020} & - & 0.121 \\
\bottomrule
\end{tabular}

\hspace{3em}

\begin{tabular}{lc}
\multicolumn{2}{c}{(d) Abbreviation details} \\
\toprule
Description & Abbreviation \\
\midrule
Utilizing distant dataset for domain tuning & \textsc{Twd}  \\
Additional tokenization pre-processing & \textsc{Tkn}  \\
Utilizing loss value as early stop signal & \textsc{Lt}  \\
Multi-view SVM with L1 regularization & mSVM-L1\\
Multi-view SVM with L2 regularization & mSVM-L2\\
\bottomrule
\end{tabular}
}

\caption{(a), (b) and (c) describe macro-averaged F1 scores for each of our 5 submitted models on the development (Dev) and test set (Test). (d) defines abbreviations used for the experimental settings. We bold the highest scores in our experiments. *~indicates our official submissions to the shared task. Top, median, mean, minimum (Min) and standard derivation (Std) values for OffensEval 2019 and 2020 (systems differ) are from Zampieri et al. (2019b, 2020).
}
\vspace{-1em}
\label{tab:tasks}
\end{table}

Table~\ref{tab:tasks} (a-c) shows the performance of our 5 submitted models selected based on their development set performance. Table~\ref{tab:tasks} (d) defines abbreviations for the experimental settings used in this section. 


The tuned BERT-\textsc{Twd}-\textsc{Lt} model achieves the best performance among our models both on development and test sets for Sub-task A, showing that adaptation of \texttt{BERT-Base} model on Twitter data can substantially boost the performance, and the loss tuning approach can further enhance the model performance. While surpassing official median scores, tuned BERT-\textsc{Twd}-\textsc{Lt} model lags behind the top official score by 0.5\% on F1 score, leading to rank 5 on this sub-task.
For Sub-task B, we also observe that the BERT-\textsc{Twd}-\textsc{Lt} model outperforms other models and official results on the test set, but performs worse than the mSVM-L2 + BERT-\textsc{Tkn}-\textsc{Twd} on the development set, and thus was not our official scored submission. This discrepancy in performance suggests different distributional characteristics between the two sets on Sub-task B. 
For Sub-task C, the mSVM-L2 + BERT-\textsc{Tkn}-\textsc{Twd} outperforms our other systems on test set, demonstrating that utilizing an ensemble approach can be effective. We observed that named entities are quite important on Sub-task C. While mSVM performs reasonably well at capturing named entities, when it is further combined with \textsc{BERT-Tkn-Twd} it is able to capture hidden relationships among tweet tokens. This leads to a considerable boost in F1 score. Since this model under-performed on the development set, it was not our official submission.

\textbf{Ablation analysis.}
To gain a sense of how different components aid our model at performing the classification tasks, we perform an ablation study for each sub-task. When comparing models' performance in Table~\ref{tab:tasks}, we see that among the given components, domain tuning (\textsc{Twd}) yields consistent improvements across all sub-tasks, justifying that training vanilla \texttt{Bert-Base} on task-related data improves the performance significantly. Interestingly, it does not hold for tokenization approach (\textsc{Tkn}).
while it was a significant component of the previous top system~\cite{Liu2019nuli}. Multi-view ensemble approach (mSVM) does not provide much improvement, although it achieves the best on the test set of Sub-task C. In Sub-task A and B, it is still behind the top scores. This might be because named entities play a crucial role in the identification of offense targets (i.e., Sub-task C). 

\begin{table}[]
\centering\small
\scalebox{0.9}{
\begin{tabular}{llcc}
\toprule
 & Tweet                                                                                                                                 & Prediction & Gold \\
 \midrule
 \multirow{2}{*}{\STAB{\rotatebox[origin=c]{90}{\footnotesize	 \tiny Task A}}} 
 & \textit{\underline{\small(A1)} @USER He deserve the worst of this all}                                                                          & OFF        & NOT  \\
 & \textit{\underline{\small(A2)} @USER @USER Dehumanize? He barely has a reflection of human}                                                                           & NOT        & OFF  \\
 \midrule
 \multirow{2}{*}{\STAB{\rotatebox[origin=c]{90}{\footnotesize	 \tiny Task B}}} 
 & \textit{\underline{\small(B1)} @USER @USER @USER I'm just checking ... is it actually illegal to have an ass that perfect???}                                                                                           & UNT        & TIN  \\
 & \textit{\underline{\small(B2)} Then to top it completely the f*** off my trainer was my boyfriend...}  & TIN        & UNT  \\
 \midrule
 \multirow{3}{*}{\STAB{\rotatebox[origin=c]{90}{\footnotesize	\tiny Task C}}} 
 & \textit{\underline{\small(C1)} Can't and won't f***ing deal with LIARS. Goodbye.}                                                                                     & GRP        & IND  \\
 &  \textit{\underline{\small(C2)} Honestly they're not even pretty and the music sucks....    }                                                     & GRP        & OTH  \\

 &  \textit{\underline{\small(C3)} ``The Democrats created the KKK” yeah my ni*** I’m gonna need you to pick up...}                                                     & OTH        & IND  \\
 \bottomrule
\end{tabular}
}
\vspace{-0.6em}
\caption{Examples of misclassified Tweets made by our model for each sub-task.}
\vspace{-1em}
\label{tab:er}
\end{table}

\textbf{Error analysis.}
To understand the limitation and qualities of the models, we qualitatively analyze the predictions from our best-performing models on several examples as shown in Table~\ref{tab:er}.  
By investigating the misclassified cases of each sub-task, 
we identified the following qualities of the misclassification examples.
1) \textit{Annotation issues (\underline{A1}, \underline{B1}, \underline{C1})}: There are tweets with labels that are not in line with interpretation of the annotation guidelines. For instance, cases such as \underline{B1} while contain profanity, they do not seem to be targeted toward certain group/individual. 
2) \textit{Absurdity (\underline{B2}): }  Social media texts can often be obscure. As such, comprehending the tweet is not only hard for the model but in some cases, humans also have trouble understanding it. Tweets like \underline{B2} can be considered offensive due to the profanity, but do not appear to contain a threat or insult and thus should not be considered targeted.
3) \textit{Sarcasm/Metaphor (\underline{A2}):} as also discussed in prior works \cite{macavaney2019hate}, tweets that contain high levels of sarcasm or metaphor are hard to be picked up by predictive models;
4) \textit{Multi-targets: (\underline{C3})} For cases in which multiple offense targets have been mentioned, it appears that the model has difficulty in picking up the true offense target. 


\section{Conclusions}
In this study, we investigated three English sub-tasks of OffensEval 2020: 1) Offensive language identification; 2) Detection if the language is targeted; and 3) Identification of the target. Specifically, we explored fine-tuning BERT model with different configurations for each sub-task. We also investigated an ensemble learning method, multi-view SVM (i.e., mSVM) model, and further combined it with BERT models to improve model performance. Our experiments demonstrate the efficacy of our approaches. Our ablation study revealed that adaptation of BERT model to task-specific data can significantly improve the classification results. Furthermore, we conducted an error analysis over the predicted labels and identified 4 common errors which can be good directions for future work.

\section*{Acknowledgements}
We thank Cristopher Flagg and Michael Kranzlein for their help.

\bibliographystyle{coling}
\bibliography{biblio}

\begin{thebibliography}{}

\bibitem[\protect\citename{Badjatiya \bgroup et al.\egroup
  }2017]{Badjatiya2017DeepLF}
Pinkesh Badjatiya, Shashank Gupta, Manish Gupta, and Vasudeva Varma.
\newblock 2017.
\newblock {Deep Learning for Hate Speech Detection in Tweets}.
\newblock In {\em Proceedings of the 26th International Conference on World
  Wide Web Companion}.

\bibitem[\protect\citename{Bird \bgroup et al.\egroup }2009]{bird2009natural}
Steven Bird, Ewan Klein, and Edward Loper.
\newblock 2009.
\newblock {\em {Natural language processing with Python: analyzing text with
  the natural language toolkit}}.
\newblock " O'Reilly Media, Inc.".

\bibitem[\protect\citename{Chen \bgroup et al.\egroup }2012]{chen2012detecting}
Ying Chen, Yilu Zhou, Sencun Zhu, and Heng Xu.
\newblock 2012.
\newblock {Detecting offensive language in social media to protect adolescent
  online safety}.
\newblock In {\em 2012 International Conference on Privacy, Security, Risk and
  Trust and 2012 International Confernece on Social Computing}. IEEE.

\bibitem[\protect\citename{Davidson \bgroup et al.\egroup
  }2017]{davidson2017automated}
Thomas Davidson, Dana Warmsley, Michael Macy, and Ingmar Weber.
\newblock 2017.
\newblock {Automated hate speech detection and the problem of offensive
  language}.
\newblock In {\em Eleventh international aaai conference on web and social
  media}.

\bibitem[\protect\citename{Devlin \bgroup et al.\egroup
  }2019]{Devlin2019BERTPO}
Jacob Devlin, Ming-Wei Chang, Kenton Lee, and Kristina Toutanova.
\newblock 2019.
\newblock {BERT: Pre-training of Deep Bidirectional Transformers for Language
  Understanding}.
\newblock In {\em Proceedings of the 2019 Conference of the North {A}merican
  Chapter of the Association for Computational Linguistics: Human Language
  Technologies}.

\bibitem[\protect\citename{Gamb{\"a}ck and Sikdar}2017]{gamback2017using}
Bj{\"o}rn Gamb{\"a}ck and Utpal~Kumar Sikdar.
\newblock 2017.
\newblock {Using convolutional neural networks to classify hate-speech}.
\newblock In {\em Proceedings of the first workshop on abusive language
  online}.

\bibitem[\protect\citename{Gururangan \bgroup et al.\egroup
  }2020]{gururangan2020-dont-stop}
Suchin Gururangan, Ana Marasovic, Swabha Swayamdipta, Kyle Lo, Iz~Beltagy, Doug
  Downey, and Noah~A. Smith.
\newblock 2020.
\newblock {Don't Stop Pretraining: Adapt Language Models to Domains and Tasks}.
\newblock In {\em Proceedings of the 58th Annual Meeting of the Association for
  Computational Linguistics}.

\bibitem[\protect\citename{Han \bgroup et al.\egroup }2019]{Han2019jhan014AS}
Jiahui Han, Shengtan Wu, and Xinyu Liu.
\newblock 2019.
\newblock {jhan014 at SemEval-2019 Task 6: Identifying and Categorizing
  Offensive Language in Social Media}.
\newblock In {\em Proceedings of the 13th International Workshop on Semantic
  Evaluation}.

\bibitem[\protect\citename{Liu \bgroup et al.\egroup }2019a]{Liu2019FuzzyML}
Han Liu, Pete Burnap, Wafa Alorainy, and Matthew~L. Williams.
\newblock 2019a.
\newblock {Fuzzy Multi-task Learning for Hate Speech Type Identification}.
\newblock In {\em The World Wide Web Conference}.

\bibitem[\protect\citename{Liu \bgroup et al.\egroup }2019b]{Liu2019nuli}
Ping Liu, Wen Li, and Liang Zou.
\newblock 2019b.
\newblock {NULI at SemEval-2019 Task 6: transfer learning for offensive
  language detection using bidirectional transformers}.
\newblock In {\em Proceedings of the 13th International Workshop on Semantic
  Evaluation}.

\bibitem[\protect\citename{MacAvaney \bgroup et al.\egroup
  }2019]{macavaney2019hate}
Sean MacAvaney, Hao-Ren Yao, Eugene Yang, Katina Russell, Nazli Goharian, and
  Ophir Frieder.
\newblock 2019.
\newblock {Hate speech detection: Challenges and solutions}.
\newblock {\em PloS one}, 14(8).

\bibitem[\protect\citename{Malmasi and Zampieri}2018]{malmasi2018challenges}
Shervin Malmasi and Marcos Zampieri.
\newblock 2018.
\newblock {Challenges in discriminating profanity from hate speech}.
\newblock {\em Journal of Experimental \& Theoretical Artificial Intelligence},
  30(2):187--202.

\bibitem[\protect\citename{Nobata \bgroup et al.\egroup
  }2016]{Nobata2016AbusiveLD}
Chikashi Nobata, Joel~R. Tetreault, Achint~Oommen Thomas, Yashar Mehdad, and
  Yi~Chang.
\newblock 2016.
\newblock {Abusive Language Detection in Online User Content}.
\newblock In {\em Proceedings of the 25th International Conference on World
  Wide Web}.

\bibitem[\protect\citename{Park and Fung}2017]{Park2017OnestepAT}
Ji~Ho Park and Pascale Fung.
\newblock 2017.
\newblock {One-step and Two-step Classification for Abusive Language Detection
  on Twitter}.
\newblock In {\em Proceedings of the First Workshop on Abusive Language
  Online}.

\bibitem[\protect\citename{Pitsilis \bgroup et al.\egroup
  }2018]{pitsilis2018detecting}
Georgios~K Pitsilis, Heri Ramampiaro, and Helge Langseth.
\newblock 2018.
\newblock {Detecting offensive language in tweets using deep learning}.
\newblock {\em arXiv preprint arXiv:1801.04433}.

\bibitem[\protect\citename{Qian \bgroup et al.\egroup
  }2018]{Qian2018LeveragingIA}
Jing Qian, Mai ElSherief, Elizabeth~M. Belding-Royer, and William~Yang Wang.
\newblock 2018.
\newblock {Leveraging Intra-User and Inter-User Representation Learning for
  Automated Hate Speech Detection}.
\newblock In {\em Proceedings of the 2018 Conference of the North {A}merican
  Chapter of the Association for Computational Linguistics: Human Language
  Technologies}.

\bibitem[\protect\citename{Rizos \bgroup et al.\egroup
  }2019]{Rizos2019AugmentTP}
Georgios Rizos, Konstantin Hemker, and Bj{\"o}rn~W. Schuller.
\newblock 2019.
\newblock {Augment to Prevent: Short-Text Data Augmentation in Deep Learning
  for Hate-Speech Classification}.
\newblock In {\em Proceedings of the 28th ACM International Conference on
  Information and Knowledge Management}.

\bibitem[\protect\citename{Rosenthal \bgroup et al.\egroup
  }2020]{rosenthal2020largescale}
Sara Rosenthal, Pepa Atanasova, Georgi Karadzhov, Marcos Zampieri, and Preslav
  Nakov.
\newblock 2020.
\newblock {A Large-Scale Semi-Supervised Dataset for Offensive Language
  Identification}.
\newblock {\em arXiv preprint arXiv:2004.14454}.

\bibitem[\protect\citename{Schmidt and Wiegand}2017]{schmidt2017survey}
Anna Schmidt and Michael Wiegand.
\newblock 2017.
\newblock {A survey on hate speech detection using natural language
  processing}.
\newblock In {\em Proceedings of the Fifth International Workshop on Natural
  Language Processing for Social Media}.

\bibitem[\protect\citename{Warner and Hirschberg}2012]{warner2012detecting}
William Warner and Julia Hirschberg.
\newblock 2012.
\newblock {Detecting hate speech on the world wide web}.
\newblock In {\em Proceedings of the second workshop on language in social
  media}.

\bibitem[\protect\citename{Waseem and Hovy}2016]{Waseem2016HatefulSO}
Zeerak Waseem and Dirk Hovy.
\newblock 2016.
\newblock {Hateful Symbols or Hateful People? Predictive Features for Hate
  Speech Detection on Twitter}.
\newblock In {\em Proceedings of the {NAACL} Student Research Workshop}.

\bibitem[\protect\citename{Wolf \bgroup et al.\egroup
  }2019]{Wolf2019HuggingFacesTS}
Thomas Wolf, Lysandre Debut, Victor Sanh, Julien Chaumond, Clement Delangue,
  Anthony Moi, Pierric Cistac, Tim Rault, R'emi Louf, Morgan Funtowicz, and
  Jamie Brew.
\newblock 2019.
\newblock {HuggingFace's Transformers: State-of-the-art Natural Language
  Processing}.
\newblock {\em arXiv preprint arXiv:1910.03771}.

\bibitem[\protect\citename{Xiang \bgroup et al.\egroup
  }2012]{xiang2012detecting}
Guang Xiang, Bin Fan, Ling Wang, Jason Hong, and Carolyn Rose.
\newblock 2012.
\newblock {Detecting offensive tweets via topical feature discovery over a
  large scale twitter corpus}.
\newblock In {\em Proceedings of the 21st ACM international conference on
  Information and knowledge management}.

\bibitem[\protect\citename{Yao}2019]{10.1145/3308560.3314196}
Mengfan Yao.
\newblock 2019.
\newblock {Robust Detection of Cyberbullying in Social Media}.
\newblock In {\em Companion Proceedings of The 2019 World Wide Web Conference}.

\bibitem[\protect\citename{Yin \bgroup et al.\egroup }2009]{Yin2009DetectionOH}
Dawei Yin, Zhenzhen Xue, Liangjie Hong, Brian~D Davison, April Kontostathis,
  and Lynne Edwards.
\newblock 2009.
\newblock Detection of harassment on web 2.0.
\newblock {\em Proceedings of the Content Analysis in the WEB}, 2:1--7.

\bibitem[\protect\citename{Zampieri \bgroup et al.\egroup
  }2019a]{Zampieri2019PredictingTT}
Marcos Zampieri, Shervin Malmasi, Preslav Nakov, Sara Rosenthal, Noura Farra,
  and Ritesh Kumar.
\newblock 2019a.
\newblock {Predicting the Type and Target of Offensive Posts in Social Media}.
\newblock In {\em Proceedings of the 2019 Conference of the North {A}merican
  Chapter of the Association for Computational Linguistics: Human Language
  Technologies}.

\bibitem[\protect\citename{Zampieri \bgroup et al.\egroup
  }2019b]{Zampieri2019SemEval2019T6}
Marcos Zampieri, Shervin Malmasi, Preslav Nakov, Sara Rosenthal, Noura Farra,
  and Ritesh Kumar.
\newblock 2019b.
\newblock {SemEval-2019 Task 6: Identifying and Categorizing Offensive Language
  in Social Media (OffensEval)}.
\newblock In {\em Proceedings of the 13th International Workshop on Semantic
  Evaluation}.

\bibitem[\protect\citename{Zampieri \bgroup et al.\egroup
  }2020]{zampieri-etal-2020-semeval}
Marcos Zampieri, Preslav Nakov, Sara Rosenthal, Pepa Atanasova, Georgi
  Karadzhov, Hamdy Mubarak, Leon Derczynski, Zeses Pitenis, and
  \c{C}a\u{g}r{\i} \c{C}\"{o}ltekin.
\newblock 2020.
\newblock {SemEval-2020 Task 12: Multilingual Offensive Language Identification
  in Social Media (OffensEval 2020)}.
\newblock In {\em arXiv preprint arXiv:2006.07235}.

\bibitem[\protect\citename{Zeng \bgroup et al.\egroup }2014]{zeng2014relation}
Daojian Zeng, Kang Liu, Siwei Lai, Guangyou Zhou, and Jun Zhao.
\newblock 2014.
\newblock {Relation classification via convolutional deep neural network}.
\newblock In {\em Proceedings of COLING 2014, the 25th International Conference
  on Computational Linguistics}.

\bibitem[\protect\citename{Zhang \bgroup et al.\egroup
  }2018]{Zhang2018DetectingHS}
Ziqi Zhang, David Robinson, and Jonathan~A. Tepper.
\newblock 2018.
\newblock {Detecting Hate Speech on Twitter Using a Convolution-GRU Based Deep
  Neural Network}.
\newblock In {\em ESWC}.

\bibitem[\protect\citename{Zhao \bgroup et al.\egroup }2017]{zhao2017multi}
Jing Zhao, Xijiong Xie, Xin Xu, and Shiliang Sun.
\newblock 2017.
\newblock {Multi-view learning overview: Recent progress and new challenges}.
\newblock {\em Information Fusion}, 38:43--54.

\end{thebibliography}

\end{document}